\documentclass[conference]{IEEEtran}
\IEEEoverridecommandlockouts

\usepackage{cite}
\usepackage{amsmath,amssymb,amsfonts}
\usepackage{algorithmic}
\usepackage{graphicx}
\usepackage{textcomp}
\usepackage{booktabs} 
\usepackage{xcolor}
\def\BibTeX{{\rm B\kern-.05em{\sc i\kern-.025em b}\kern-.08em
    T\kern-.1667em\lower.7ex\hbox{E}\kern-.125emX}}
\begin{document}

\title{What Matters, When? Diagnosing and Improving Conditional Visual Grounding in Visuomotor Imitation Policies\\

}

\author{
\begin{tabular}{c}

Vivek Chavan$^{1,2,\ast}$ \quad
Pengtao Xie$^{2,\ast,\dagger}$ \quad
Yahuan Shi$^{2,\dagger}$ \quad
Oliver Heimann$^{1}$ \quad
Kevin Haninger$^{1,\ddagger}$ \quad
Jörg Krüger$^{1,2,\ddagger}$

\\[6pt]

$^{1}$\textit{Fraunhofer Institute for Production Systems and Design Technology IPK}

\\

$^{2}$\textit{Technische Universität Berlin}

\\[3pt]

{\footnotesize
$^{\ast}$Equal contribution.
\qquad
$^{\dagger}$Work conducted as part of student projects at Fraunhofer IPK.
\qquad
$^{\ddagger}$Supervision
}

\end{tabular}
}

\maketitle

\begin{abstract}
Visuomotor imitation policies can achieve high performance in curated environments yet fail when visually similar objects compete with task-relevant entities. We study this behavior as a problem of \emph{conditional visual grounding}: which visual entity matters depends on the current manipulation phase and, in more complex tasks, on the inferred task state. Using Action Chunking with Transformers (ACT), we systematically vary object and receptacle competitors and localize failures to picking and placement. The resulting errors are cue- and phase-specific and are accompanied by corresponding changes in learned visual representations. Guided by this diagnosis, we evaluate distractor augmentation, phase-dependent attention regularization, and appearance-based visual prompting as complementary interventions for strengthening task-relevant grounding while preserving spatial information required for control. These interventions substantially improve robustness in simulation and on a physical UR3e. We further examine the same diagnosis-intervention principle in a pretrained vision-language-action policy on a state-conditioned instrument-handling task, where the correct destination depends on the observed instrument state. Together, the results support conditional visual grounding as a useful framework for diagnosing and improving robustness across distinct visuomotor policy-learning regimes. 8 Page paper!
\end{abstract}


\section{Introduction}
\label{sec:introduction}

Vision-based imitation policies are commonly trained and evaluated under curated visual conditions, while deployment scenes can contain visually similar objects, receptacles, and irrelevant clutter~\cite{xie2024generalization,dong2025imitdiff}. Adding such competitors changes neither the manipulation objective nor the underlying robot dynamics, yet can substantially reduce policy success~\cite{chen2025causalact,dong2025imitdiff}. Aggregate success alone, however, cannot distinguish between a policy that has lost the ability to execute the manipulation and one that retains the motor behavior but applies it to the wrong entity.

We study this distinction as \emph{conditional visual grounding}. Given candidate referents $\mathcal{R}_t$, an instruction $\ell$, and execution context $c_t$, correct action requires selecting a context-dependent referent $r_t^\star \in \mathcal{R}_t$. Crucially, the relevant referent is not necessarily static. For Action Chunking with Transformers (ACT)~\cite{ref1}, we examine phase-conditioned relevance: the manipulated object is critical during target acquisition and grasping, whereas the receptacle becomes critical during placement. In our vision-language-action (VLA) case study, relevance is additionally state-conditioned: the observed state of a medical instrument determines which receptacle is the correct destination.

We ask whether visual failures can be localized to the cue, referent, and execution stage at which grounding becomes incorrect, and whether this diagnosis suggests effective interventions. Controlled ACT experiments first expose distinct object- and receptacle-selection bottlenecks. We then evaluate lightweight object-centric interventions and validate their effect in simulation and on a physical UR3e. Finally, we examine whether the same diagnosis-intervention principle remains useful in a pretrained VLA policy on a state-conditioned routing task.

\section{Related Work}
\label{sec:related}

Visuomotor imitation policies are vulnerable to spurious observation-action correlations~\cite{dehaan2019causal}, and recent work has systematically quantified generalization failures under changes in object appearance, scene configuration, and visual distractors~\cite{xie2024generalization}. In particular, Causal-ACT shows that ACT can exploit irrelevant distractor correlations~\cite{chen2025causalact}, while ImitDiff and DRAIL improve robustness through semantic guidance and task-relevant region-aware representations, respectively~\cite{dong2025imitdiff,hattori2026drail}.

Complementary approaches provide explicit spatial or visual guidance through grounding masks and visual prompts~\cite{huang2025roboground,muttaqien2025visualprompting}. Visual robustness is likewise unresolved in pretrained vision-language-action policies: BYOVLA demonstrates sensitivity to task-irrelevant visual content and mitigates it through run-time observation interventions~\cite{hancock2024runtime}, while recent work explicitly supervises task-relevant factors in VLA action representations~\cite{jia2026guidedvla}. Our focus is complementary; we jointly diagnose \emph{which} visual cues induce incorrect grounding, \emph{which} task referent is affected, and \emph{when} the resulting error becomes behaviorally consequential, and use this diagnosis to motivate interventions across distinct visuomotor policy-learning regimes.

\section{Method}
\label{sec:method}

\paragraph{Controlled diagnosis with ACT.} We study two simulated pick-and-place tasks designed to separate target and destination grounding. In Task~1, a randomly placed object is moved to a fixed receptacle; in Task~2, both object and receptacle positions are randomized, requiring visual localization during both picking and placement. ACT is trained from 100 clean scripted demonstrations per task. At evaluation, held-out object and receptacle competitors match the target in color, shape, or neither, with one to three competitors per class. The \emph{full mixed} condition contains one competitor from each object and receptacle class. In addition to end-to-end success, we measure $P(\mathrm{pick})$, $P(\mathrm{lift}\mid\mathrm{pick})$, and $P(\mathrm{place}\mid\mathrm{pick},\mathrm{lift})$, allowing failures to be localized within the manipulation sequence. Simulation results aggregate three training seeds, four evaluation seeds, and 50 closed-loop rollouts per seed.

\begin{figure*}[t]
\centering
\includegraphics[width=0.85\textwidth]{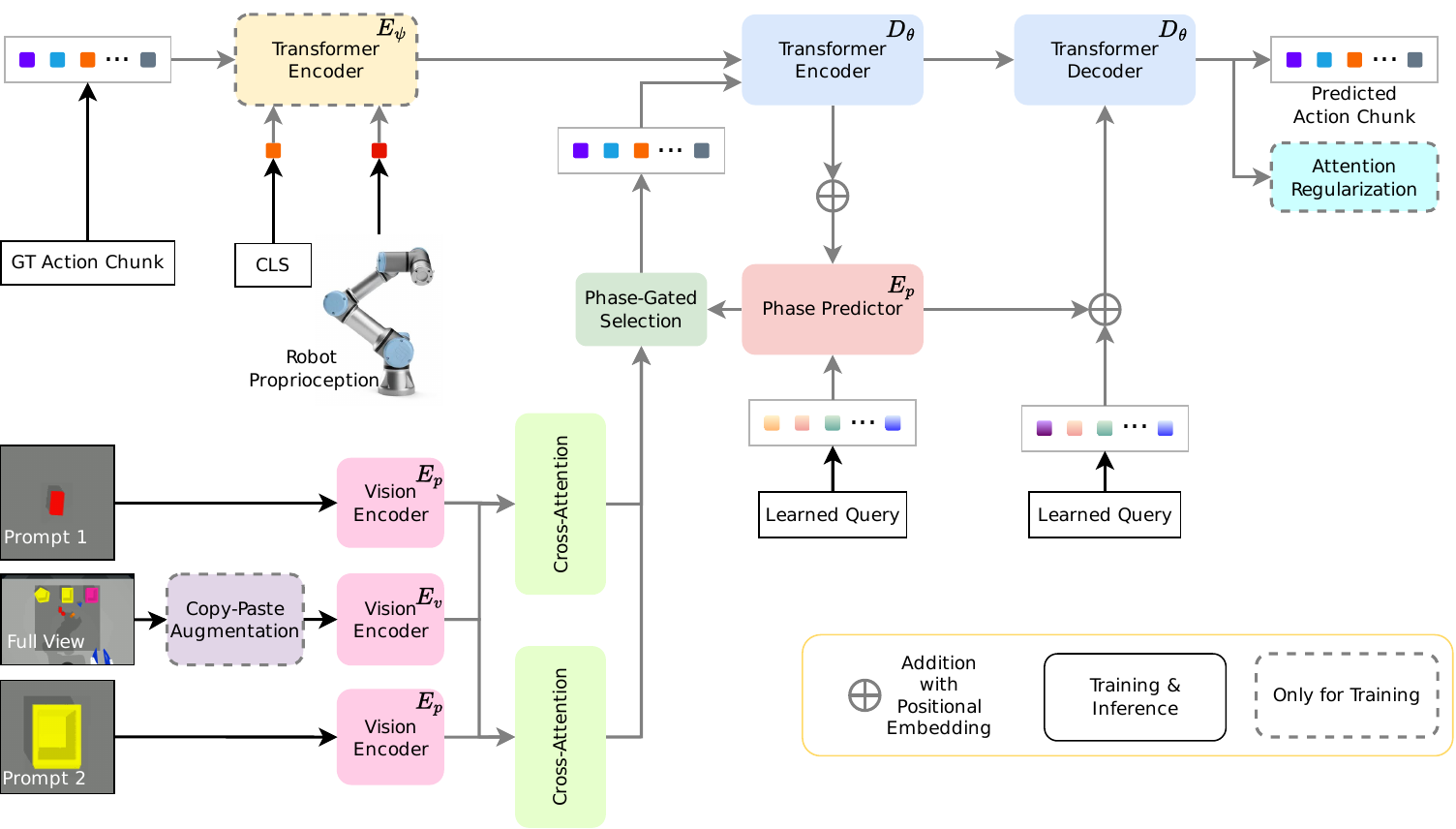}
\caption{\textbf{ACT-Modified architecture.} The full camera observation and phase-selected visual prompt are encoded into the visual memory consumed by the ACT transformer decoder. A phase predictor selects the object or receptacle prompt according to the current manipulation phase, while robot proprioception (\texttt{qpos}) provides state conditioning. The decoder combines visual memory, learned queries, cross-attention, and positional embeddings to predict an action sequence. The action-sequence encoder and \texttt{CLS} latent path form the CVAE posterior during training only; during inference, future actions are unavailable and the latent is fixed to the prior mean $z=0$.}
\label{fig:act-modified-architecture}
\end{figure*}

\paragraph{Diagnosis-driven intervention.} The diagnosis motivates three complementary mechanisms targeting incorrect grounding. Image-space copy-paste augmentation inserts synthetic competitors into otherwise clean demonstrations, reducing the reliability of incidental appearance correlations. Phase-dependent attention regularization supervises one decoder cross-attention head toward the currently relevant object or receptacle and away from synthetic distractors. Appearance-based visual prompting provides positionless crops of the target object and receptacle, with a jointly trained phase predictor selecting the appropriate prompt during execution. Figure~\ref{fig:act-modified-architecture} summarizes the resulting ACT-Modified architecture.

Let $A$ denote the supervised head's normalized attention, $\mathcal{L}_{\phi}$ the phase-prediction loss, $M_{\phi}$ the phase-relevant mask, and $M_d$ the distractor mask. Training uses
\begin{equation}
\mathcal{L}=\mathcal{L}_{\mathrm{ACT}}+\lambda_{\phi}\mathcal{L}_{\phi}+\lambda_{+}\!\left(1-\langle A,M_{\phi}\rangle\right)+\lambda_{-}\langle A,M_d\rangle .
\label{eq:objective}
\end{equation}
The visual crop specifies target appearance but contains no explicit full-frame coordinates. Simulation crops and masks use privileged segmentation, while hardware prompts are obtained from an initial manual annotation followed by SAM~2 segmentation~\cite{ravi2024sam2}. Because prompt-bearing policies receive additional target information, augmentation-only and augmentation-plus-attention variants are evaluated separately to isolate robustness gains that do not depend on explicit target specification.

\paragraph{Pretrained-VLA case study.} To examine whether the same grounding perspective remains useful beyond task-specific ACT policies, we fine-tune $\pi_{0.5}$~\cite{intelligence2025pi05} on 231 teleoperated episodes of a seven-step instrument-handling procedure observed through fixed-base and wrist cameras. Five subgoals are evaluated, including two spatially ambiguous routing subgoals in which the correct destination depends on the observed state of the medical instrument. We compare a prompt-trained checkpoint with and without its expected RGB cue and a regularized checkpoint evaluated on clean RGB input. Five trials per subgoal and condition yield 75 trials. For the routing subgoals, the prompt variants receive the generic destination instruction \textit{highlighted}, whereas the regularized variant receives explicit left/right destination text; the conditions are therefore not directly comparable.

\section{Experiments and Results}
\label{sec:results}

\paragraph{ACT failures are cue-specific and stage-localized.} Standard ACT reaches 98.5\% and 99.7\% success in clean Tasks~1 and~2, confirming that both manipulation routines are learned. Under three color-matched object competitors in Task~2, $P(\mathrm{pick})$ falls to 39.2\%, while $P(\mathrm{lift}\mid\mathrm{pick})$ and $P(\mathrm{place}\mid\mathrm{pick},\mathrm{lift})$ remain 93.8\% and 95.5\%. In contrast, two shape-matched receptacle competitors leave $P(\mathrm{pick})$ at 97.5\% but reduce conditional placement to 33.9\%. Within the evaluated assets, object selection is therefore most sensitive to color, whereas receptacle selection is most sensitive to shape. The effect increases with competitor count and compounds when both bottlenecks are active: under full mixed distractors, end-to-end success falls to 39.5\% on Task~1 and 14.0\% on Task~2. The preservation of downstream conditional success after correct selection indicates that much of the degradation arises from incorrect grounding rather than loss of the learned manipulation routine.

\paragraph{Object-centric interventions recover robustness.} Table~\ref{tab:act-results} summarizes the central behavioral results under full mixed distractors. Augmentation alone reaches 100.0\% on Task~1 and 64.0\% on Task~2 without explicit target prompts, showing that substantial robustness can be recovered by disrupting spurious appearance correlations during training. Its remaining Task~2 deficit is concentrated in randomized-receptacle localization. ACT-Modified reaches 94.5\% and 88.5\% on Tasks~1 and~2, respectively.

\begin{table}[t]
\centering
\small
\caption{\textbf{End-to-end success under full mixed distractors.} Simulation contains 600 rollouts per cell; hardware contains 20 trials per cell. T1 uses a fixed receptacle and T2 a randomized receptacle.}
\label{tab:act-results}
\begin{tabular}{@{}lcccc@{}}
\toprule
& \multicolumn{2}{c}{Simulation (\%)} & \multicolumn{2}{c}{UR3e (\%)} \\
Method & T1 & T2 & T1 & T2 \\
\midrule
Standard ACT      & 39.5  & 14.0 & 0.0  & 0.0 \\
Augmentation only & 100.0 & 64.0 & -   & -  \\
ACT-Modified      & 94.5  & 88.5 & 65.0 & 60.0 \\
\bottomrule
\end{tabular}
\end{table}

On the physical UR3e, standard and modified ACT remain comparable in clean scenes: 17/20 versus 17/20 successes on Task~1 and 16/20 versus 15/20 on Task~2. Under mixed distractors, standard ACT fails all tested trials, whereas ACT-Modified succeeds in 13/20 and 12/20 trials. Given the sample size, these experiments establish that the behavioral ordering observed in simulation persists on hardware rather than reproducing the complete mechanism analysis in the physical setting.

\paragraph{Robust representations require invariance without loss of task geometry.} To examine what changes inside the policy, we compare standard ACT, augmentation-only ACT, and ACT-Modified at the vision encoder, transformer memory, and decoder state. Matched clean and full-distractor activations are evaluated using cosine shift, pick/place centroid separation, and Task~2 left/right container structure.

\begin{figure*}[t]
\centering
\setlength{\tabcolsep}{2pt}
\begin{tabular}{@{}ccc@{}}
\includegraphics[width=0.315\textwidth]{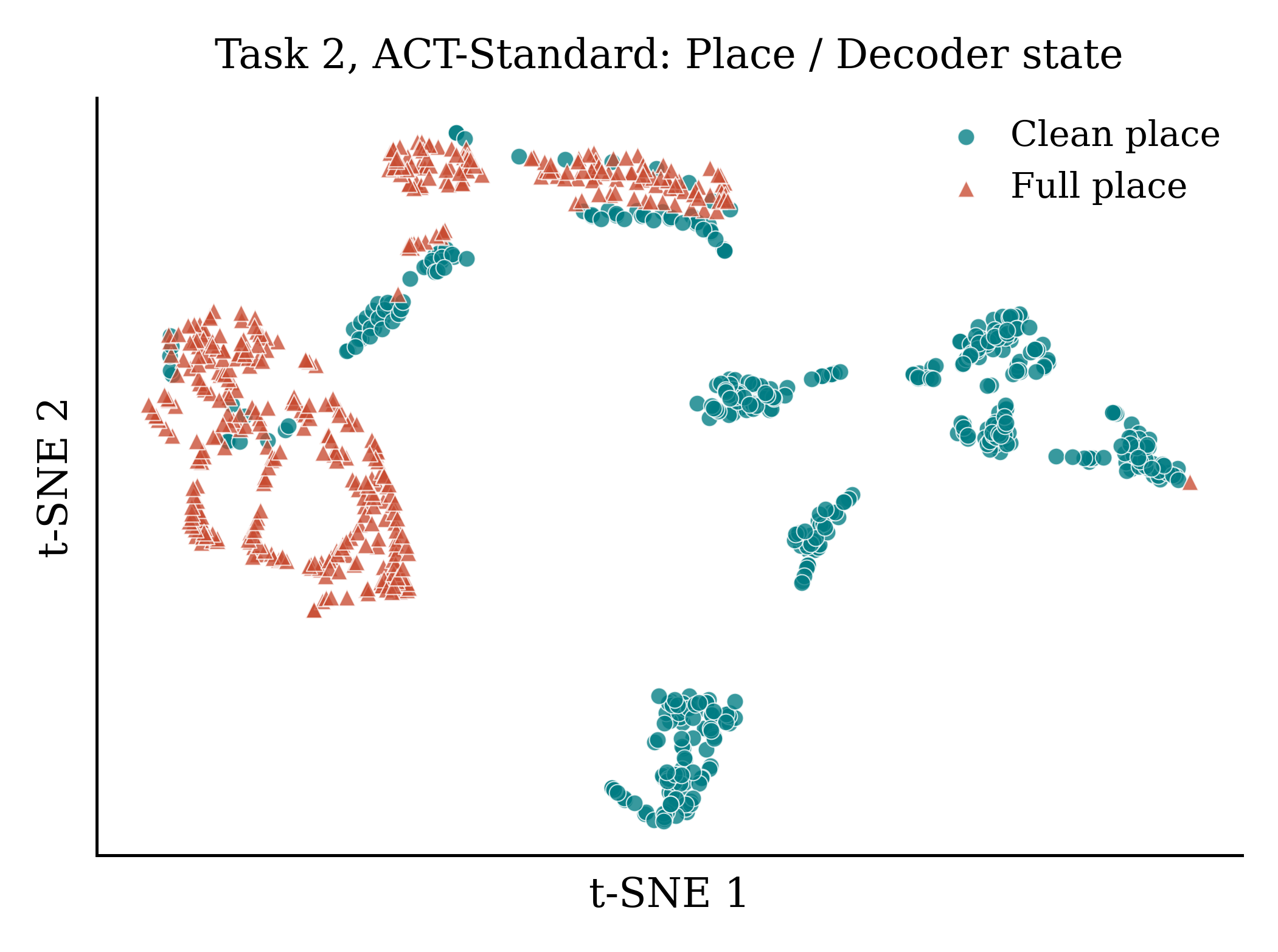} &
\includegraphics[width=0.315\textwidth]{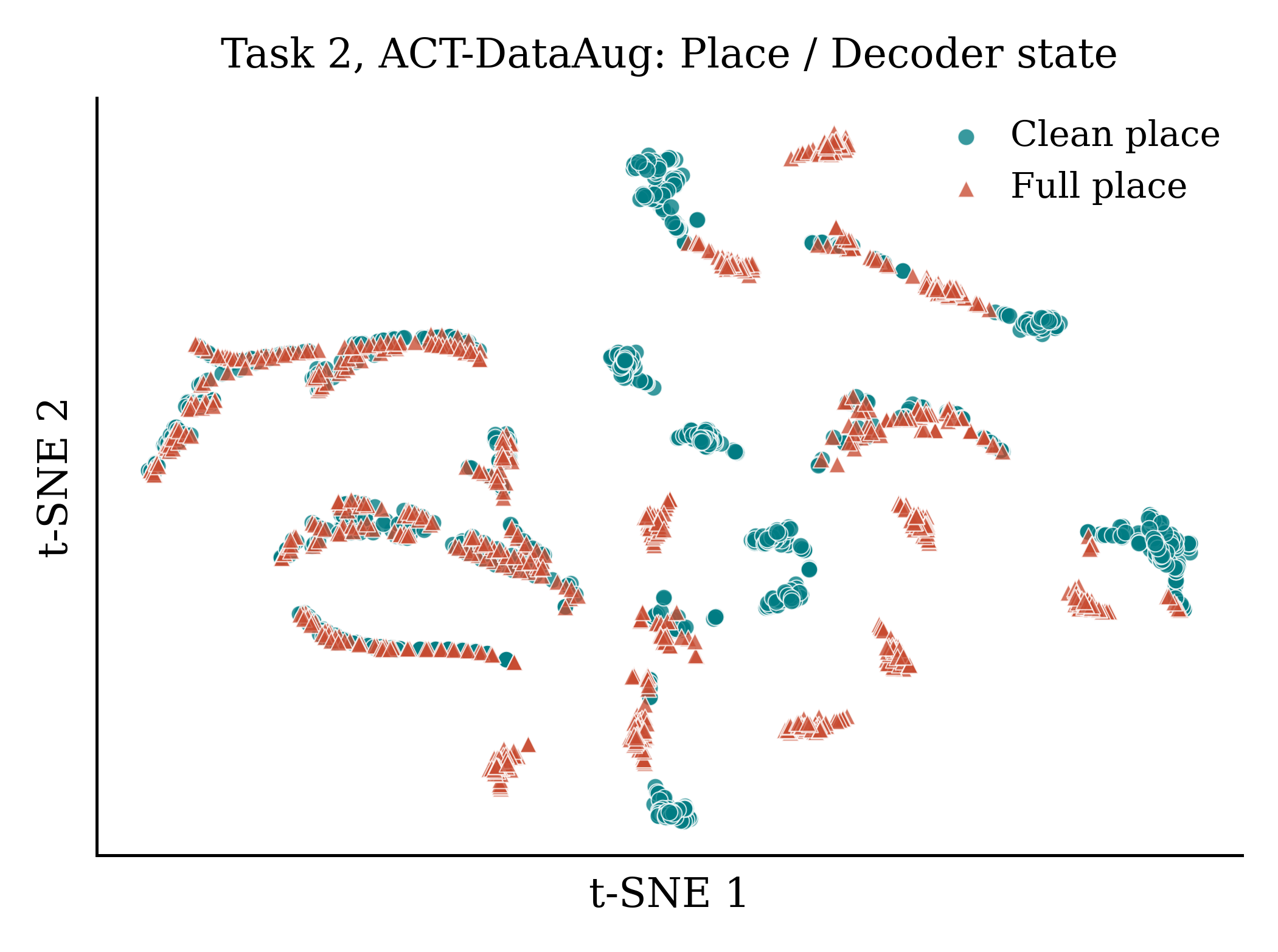} &
\includegraphics[width=0.315\textwidth]{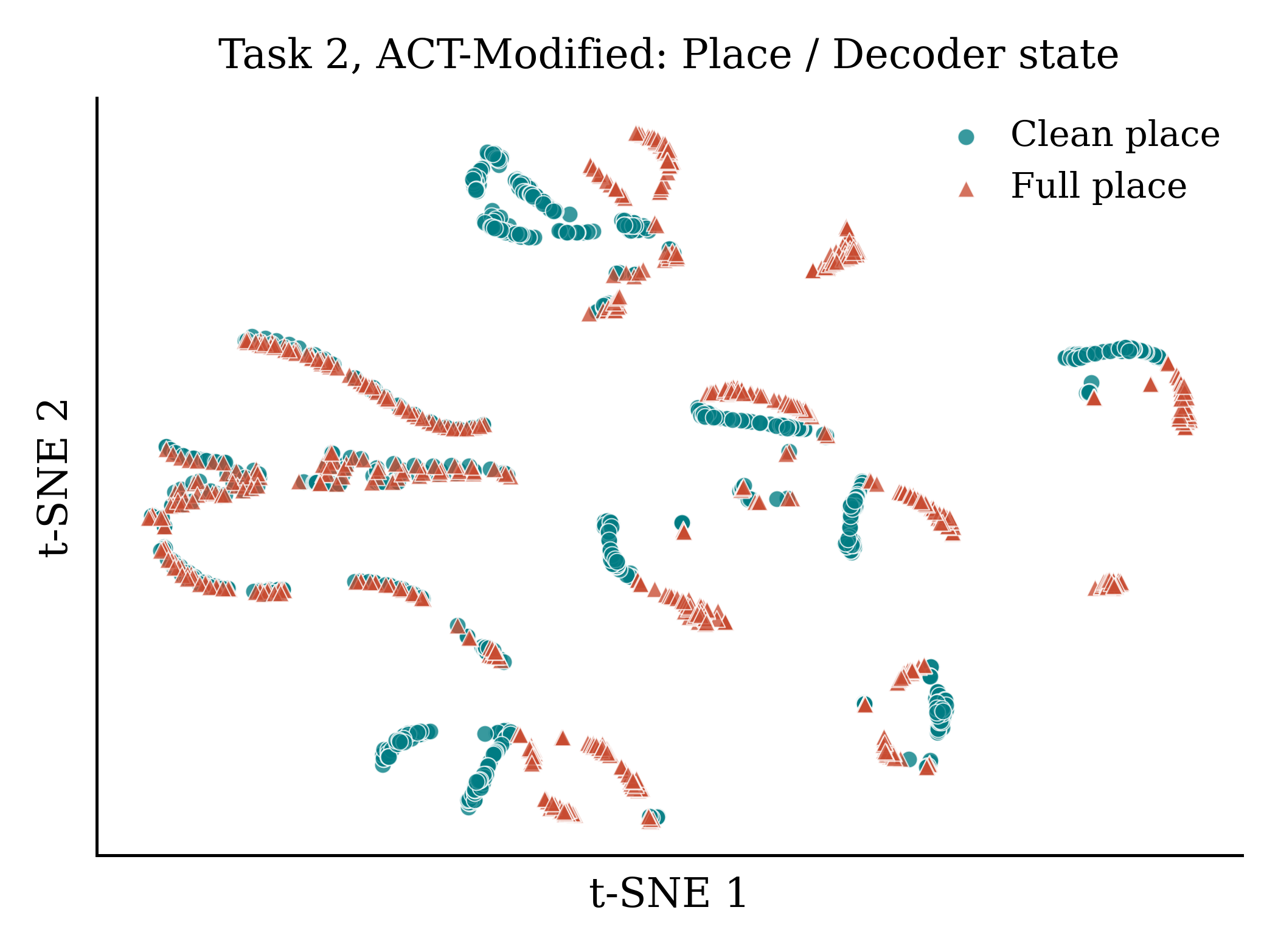} \\
{\scriptsize (a) Standard: clean/full shift} &
{\scriptsize (b) DataAug: clean/full shift} &
{\scriptsize (c) Modified: clean/full shift} \\[0.6em]
\includegraphics[width=0.315\textwidth]{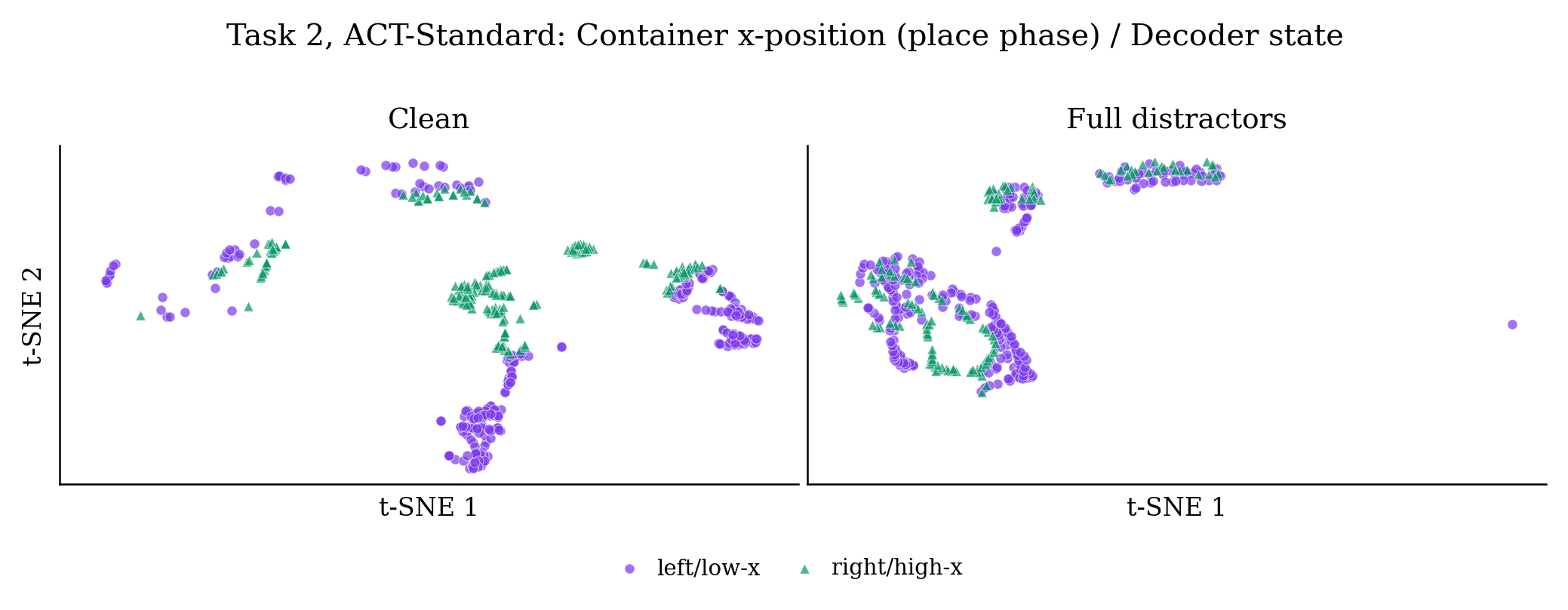} &
\includegraphics[width=0.315\textwidth]{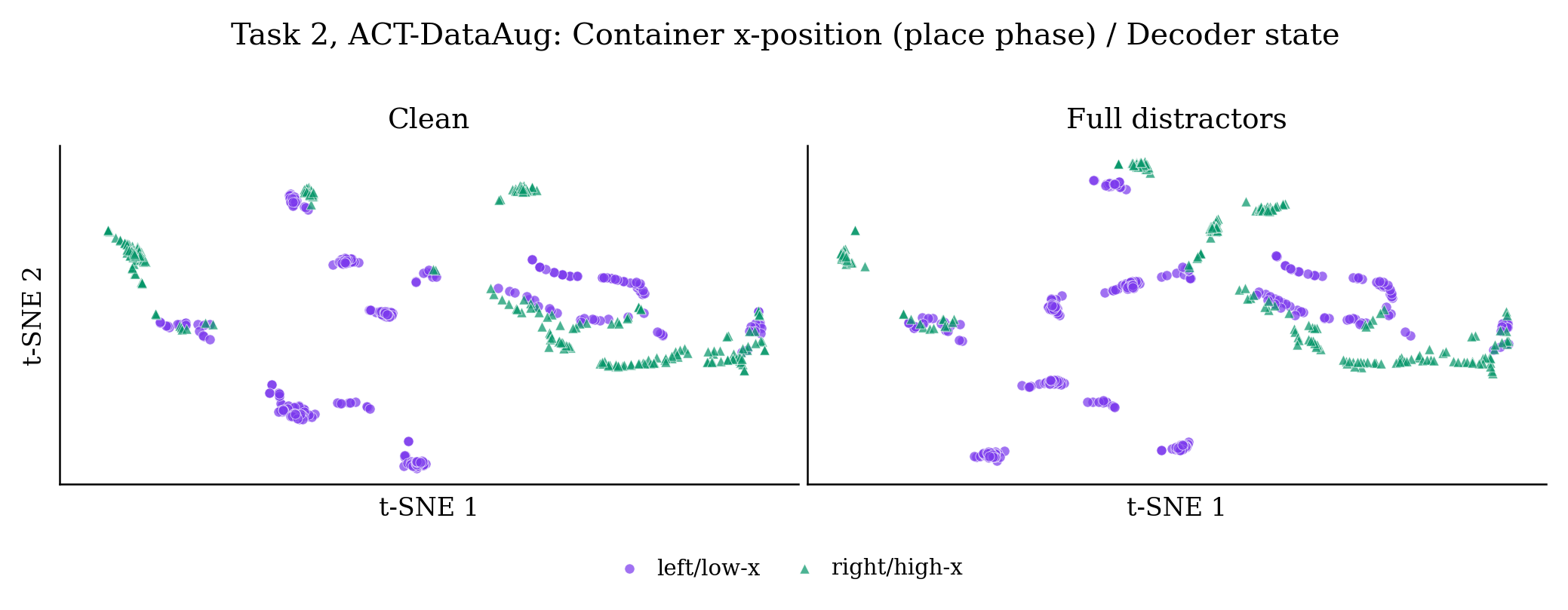} &
\includegraphics[width=0.315\textwidth]{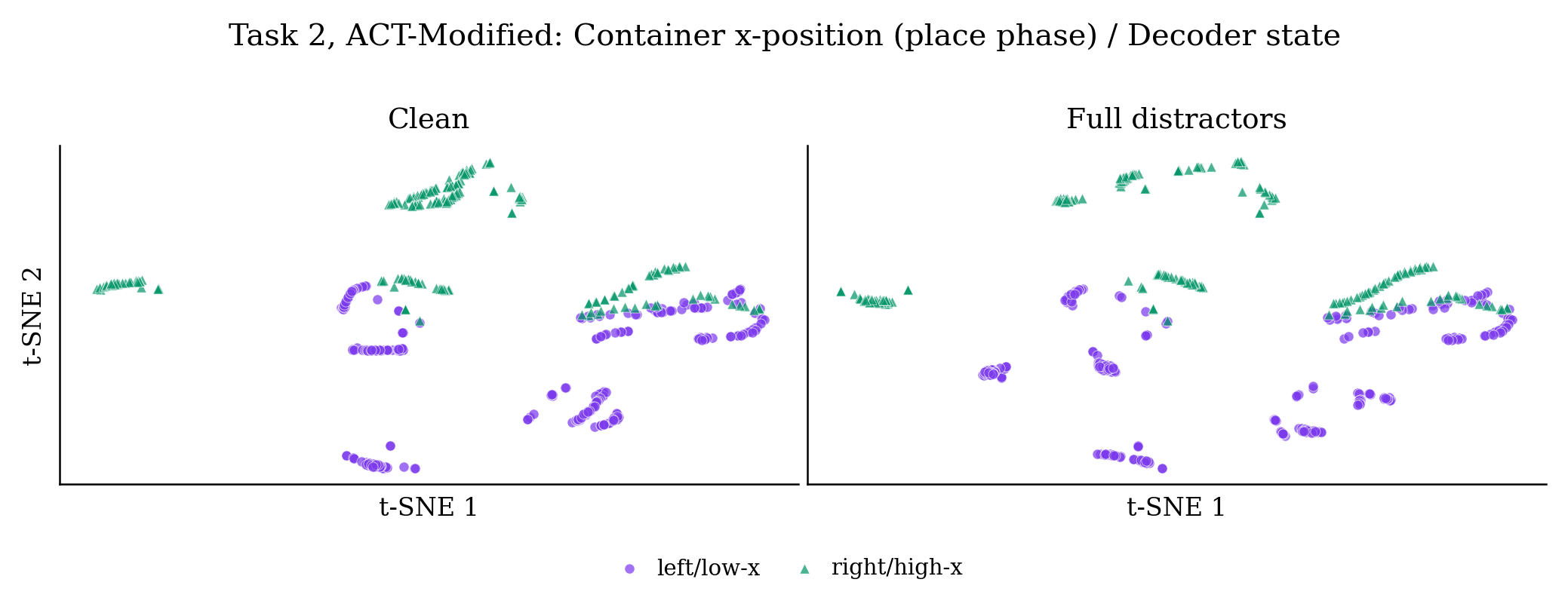} \\
{\scriptsize (d) Standard: container geometry} &
{\scriptsize (e) DataAug: container geometry} &
{\scriptsize (f) Modified: container geometry} \\
\end{tabular}
\caption{\textbf{Robust placement requires both distractor invariance and retained target geometry.} The top row compares clean and full-distractor Task~2 place-phase decoder states. The bottom row colors the corresponding representations by the target container's left/right position, with clean and full conditions shown in a shared within-model projection. Augmentation-only ACT exhibits the strongest clean/full invariance, whereas ACT-Modified preserves substantially clearer container-position structure. The t-SNE projections are illustrative; quantitative comparisons in Table~\ref{tab:ea-representation} are computed in the original feature space.}
\label{fig:ea-representation}
\end{figure*}

\begin{table}[t]
\centering
\scriptsize
\caption{\textbf{Task~2 decoder-state diagnostics.} Shift is the place-phase clean-to-full cosine distance. Phase retention measures full-to-clean pick/place centroid separation. Geometry is the full-distractor container-position silhouette.}
\label{tab:ea-representation}
\setlength{\tabcolsep}{3.0pt}
\renewcommand{\arraystretch}{1.08}
\begin{tabular}{@{}lccc@{}}
\toprule
Metric & Standard & DataAug & Modified \\
\midrule
Place shift $\downarrow$ & 0.1993 & \textbf{0.0011} & 0.0125 \\
Phase retained $\uparrow$ & 13.1\% & 94.8\% & \textbf{108.5\%} \\
Container silhouette $\uparrow$ & 0.0095 & 0.0889 & \textbf{0.2037} \\
Conditional placement $\uparrow$ & 33.3\% & 66.3\% & \textbf{100.0\%} \\
\bottomrule
\end{tabular}
\end{table}

Standard ACT's decoder state shifts substantially under distractors during placement and retains only 13.1\% of its clean pick/place separation. Both object-centric variants strongly suppress this shift and preserve phase structure. The augmentation-only model provides a critical counterexample to distractor invariance as a sufficient explanation of robustness: it is more invariant than ACT-Modified, yet retains weaker container-position geometry and reaches only 66.3\% conditional placement compared with 100.0\% for ACT-Modified. Across the three policies, destination-geometry preservation follows placement performance more closely than clean-to-distractor invariance alone. Robust behavior is therefore associated with suppressing nuisance-induced variation while preserving the spatial structure required by the currently relevant referent.

The representation analyses remain correlational, and attention maps are likewise treated as diagnostic rather than causal explanations. We therefore complement them with a direct prompt intervention. In a separate cup-selection probe, replacing the target prompt with a distractor prompt increases distractor selection from 0-1\% to 35-91\%. Because the visual crop contains target appearance but no explicit full-frame coordinates, this intervention shows that prompt identity can actively redirect closed-loop target selection rather than merely correlate with successful behavior.

\paragraph{The VLA case study provides preliminary cross-regime evidence.} The prompt-trained $\pi_{0.5}$ checkpoint succeeds in 20/25 trials when evaluated without its expected RGB cue and in 25/25 when the cue is restored. The entire deficit is concentrated in the ambiguous routing subgoals: routing success increases from 5/10 to 10/10, while the remaining instrument-handling subgoals are unaffected. The regularized clean-input checkpoint also reaches 25/25 overall and 10/10 on routing. Thus, the observed deficit again localizes to selection of a context-dependent destination rather than to execution of the remaining manipulation sequence. Because the experiment is small and the guided conditions are not equally informed, it provides convergent behavioral evidence for the conditional-grounding perspective rather than evidence that ACT and $\pi_{0.5}$ share the same internal shortcut.

\section{Discussion and Outlook}
\label{sec:discussion}

The ACT and VLA studies instantiate conditional visual grounding differently. In ACT, manipulation phase determines whether the object or receptacle is relevant; in the VLA task, the observed instrument state determines the relevant destination. Across both settings, competent motor behavior can coexist with incorrect selection of a context-dependent referent. The contribution is therefore not that both architectures fail identically, but that analyzing \emph{what} must be grounded \emph{when} provides a common framework for localizing visual failures and designing targeted interventions. The ACT representation analysis further shows that robustness requires not only suppressing distractor-induced variation but also preserving the task-relevant geometry required for control.

Several limitations bound these conclusions. The observed ACT color-shape hierarchy is specific to the evaluated assets and requires validation with fully counterbalanced visual factors. Prompt-bearing ACT variants receive additional target information, although augmentation-only results demonstrate substantial robustness gains without this advantage. Attention and representation analyses remain correlational, the hardware evaluation contains only 20 trials per cell, and the exploratory VLA study contains only five trials per subgoal with non-equivalent guidance conditions. The VLA task additionally uses a visible contamination surrogate and fixed routing rule rather than general contamination understanding.

\bibliographystyle{IEEEtran}
\bibliography{main}

\end{document}